\documentclass{article}

\usepackage[nonatbib]{nips_2017}
\pdfoutput=1 
\usepackage{nips_2017}
\usepackage{subcaption}
\usepackage{cellspace}
\newcommand\Tstrut{\rule{0pt}{2.6ex}}         
\newcommand\Bstrut{\rule[-0.9ex]{0pt}{0pt}}
\usepackage[utf8]{inputenc} 
\usepackage[T1]{fontenc}    
\usepackage{hyperref}       
\usepackage{url}            
\usepackage{booktabs}       
\usepackage{amsfonts}       
\usepackage{nicefrac}       
\usepackage{microtype}      

\title{Clustering and Learning from Imbalanced Data}

\author{
  Naman Deep Singh \\
   \And
  Abhinav Dhall \\
}

\begin{document}

\maketitle

\begin{abstract}
   A learning classifier must outperform a trivial solution, in case of imbalanced data, this condition usually does not hold true. To overcome this problem, we propose a novel data level resampling method - Clustering Based Oversampling for improved learning from class imbalanced datasets.
  The essential idea behind the proposed method is to use the distance between
  a minority class sample and its respective cluster centroid to infer the number
  of new sample points to be generated for that minority class sample. The proposed algorithm has very less dependence on the technique used for finding
  cluster centroids and does not effect the majority class learning in any way. It also improves learning from imbalanced data by incorporating the distribution structure of minority class samples in generation of new data samples.
  The newly generated minority class data is handled in a way as to prevent
  outlier production and overfitting. Implementation analysis on different datasets using deep neural networks as the learning classifier
  shows the effectiveness of this method as compared to other synthetic data
  resampling techniques across several evaluation metrics.
\end{abstract}

\section{Introduction}
\label{intro}
The amount of data generated is increasing every day, this also increases the demand for  learning systems which can predict, classify and analyse the data efficiently. Classification is the type of problem most commonly solved by predictive systems, it can be of two types: binary classification \cite{mccormick2012dynamic} and multi-class classification \cite{wu2004probability,sun2006boosting}.  When a general classifier encounters imbalanced data, it favors the majority class sample. Imbalanced data is a scenario when the number of instances of one class are scanty in comparison to other classes. This causes classical classifier systems to neglect minority class instances and emphasize on majority class, resulting in a skewed classification accuracy. This accuracy might be high but minority class is misclassified. Imbalance problem can either be multi class or binary class classification. Most of the multi class imbalance problems \cite{har2003constraint} are generally converted to binary class imbalance problem (using binarization etc.) and then solved. This work is carried out for binary class imbalance problem.\\
Class imbalance problem has its applications in many real world problems. It is imperative in certain cases to correctly classify the minority class correctly. Finding fraudlent cases in health insurance,  automobile insurance fraud detection \cite{1277822}, credit card fraud detections  \cite{chan1998toward} and other cases related to financial sector are most common. The need of better classification in imbalance problem also arises while classsifying medical datasets (\cite{bhattacharya2017icu}). In cancer detection positive class cases can be very rare (1:100), sometimes even more adverse \cite{weiss2004mining,pearson2003imbalanced} and detection of these cases is very important for obvious reasons. Improper classification in such cases can have serious repercussions.\\
Nowadays, researchers working on this problem have emphasized on creating learning classifiers that are better suited to handle class imbalance \cite{Zhang2015} and less attention is being given to data level imbalance rectification techniques.
All the data level resampling techniques fall under either oversampling, which regenerates data of the minority class to make it in commensurate measures to the  majority class through some algorithms; or undersampling, which resamples the majority class to level it with the instances of the minority class. As mentioned, imbalance class problem is also present in multi-class classification problems \cite{sun2006boosting,abe2004iterative} but the proposed method has been showcased only for the bi-class imbalance classification. 
At algorithmic level, cost enhanced versions of learning algorithms like SVM, neural networks have been used but to some fruition. These cost sensitive methods are known to make learning classifiers more susceptible to variations in minority class samples.\\
In our proposed method, we have incorporated the randomness factor of random sampling and nearness approach of synthetic sampling technique for clustering and generation of new samples. This combination ensures reduction of overfitting and better performance. 
State of the art methods suffer from two problems. The first is the curse of dimensionality, it has been seen that methods like SMOTE \cite{chawla2002smote} and the one's built on top of it lag in performance when applied to high dimensional data. Secondly, these methods in one way or the other interfere with the data distributions of majority class space. The proposed method, CBOS (Clustering based oversampling) has been conceived to tackle these two fundamental problems. CBOS has been compared with the most commonly used oversampling techniques like SMOTE \cite{chawla2002smote}, SMOTE-ENN \cite{batista2004study} and ADASYN \cite{he2008adasyn}. We have used several datasets and used deep neural networks as our classifier. Although existing methods show promising results, our method improves upon these methods across different performance metrics, as can be seen in experiment and discussions section.\\
The rest of the paper is organized as follows. In Section \ref{sec2}, the past work done related to the class imbalance problem is reviewed. Section \ref{sec3} gives a detailed description of the proposed method. Descriptive analysis of learning algorithms and the performance metrics used has been done in Section \ref{sec4}. Experimental study, results obtained and related discussions have been made in Section \ref{sec5}. Finally, the paper is concluded in Section \ref{sec6}.

\section{Related Work}
\label{sec2}
A lot of work on imbalance class problem is present in literature. More recent work focuses on making learning algorithms adept to learn from imbalanced distributions whereas distant literature has a lot of work on data level resampling.
Data level resampling is one of the many ways to handle class imbalance problem. As we propose a clustering based oversampling method, Lin et al. proposed a clustering based undersampling technique with Multi layer perceptron in \cite{lin2017clustering}.
Random oversampling and random undersampling are the two most basic techniques in this category. Random oversampling suffers from overfitting whereas random undersampling leads to underfitting due to loss of data. Synthetic resampling techniques\cite{chawla2002smote, mani2003knn, batista2004study} were proposed to overcome this problem. In 2002, a sampling based algorithm called SMOTE (Synthetic Minority Over-Sampling Technique) was introduced. SMOTE \cite{chawla2002smote} balances the class distribution of original data sets by incrementing some virtual samples. While generating virtual examples SMOTE does not take into account the neighboring examples from other classes, which results in an increase in class overlapping and may introduce noise. Some new techniques, which are built upon SMOTE have come up, SMOTE-ENN\cite{batista2004study}, SMOTE-Tomek \cite{batista2004study}. Adaptive Synthetic (ADASYN) \cite{he2008adasyn} is based on the idea of adaptively generating minority data samples according to their distributions using K nearest neighbor. A detailed review of all synthetic techniques is given by \cite{1755-1315-58-1-012031}. \\
Cost Sensitive learning techniques for imbalance use a cost-matrix for
different types of errors to aid learning from imbalanced data sets. It does not affect the data distribution but different cost matrices are used for misclassification of minority class samples than for majority class samples \cite{sun2007cost}. Cost-sensitive neural network models have been deeply studied in \cite{zhou2006training}. A threshold-moving technique has been used in this method to adjust the output threshold toward majority classes, such that
high-cost (minority) instances are less likely to be misclassified. Liu et al. \cite{liu2006influence} study the effects of imbalanced data on cost sensitive learning and different cost sensitive learning techniques have been compared in \cite{thai2010cost} using several imbalanced datasets. According to \cite{wang2016training}, these methods are only applicable when the specific cost values of misclassification are known. \\
Kernel based methods have also been used to study class imbalance problem. In \cite{wu2005kba}, Wu et al. propose a method to adjust the class boundary by adapting the kernel matrix in SVMs according to the distribution of the data. Ensemble models are also known to improve performance while facing the challenge of imbalance data. Bagging \cite{quinlan1996bagging} forms another class of techniques for handling imbalanced data sets. Random Forests, a form of bagging are not good to handle imbalance data as they are based on decision trees, which are adversely susceptible to class imbalance, but when used with suitable resampling technique they can be effective. An ensemble algorithm based on balanced Random Forest for buyer prediction problem is given by Yagci et al. \cite{7495927}. 
A class-wise weighted voting approach to Random Forest for class imbalance problem for medical data has been given in \cite{8246503}. Recently,  unsupervised learning \cite{NIAN201658} techniques have been used to balance data based on unsupervised spectral rankings in auto insurance frauds.\\
A new weight adjustment factor is used by Wonji et al. \cite{LEE201792} in SVMs, and the SVM has been used as a base classifier in AdaBoost algorithm. Supposedly, this method \cite{LEE201792} solves class overlapping and small disjuncts problems which are commonly seen in clssification problems. According to \cite{galar2012review}, algorithmic level and cost-sensitive approaches are more problem dependent, whereas data level and ensemble learning methods are more pervasive as they can be used in independence to the learning classifier. A comprehensive comparison of ensemble methods for class imbalance can be seen in \cite{galar2012review}. Class switching, a mechanism to produce training perturbed sets, has proven to perform well in slightly imbalanced scenarios. In \cite{GONZALEZ201712} class switching's potential to deal with highly imbalanced data has been analysed and a new ensemble approach based on switching using nearest enemy distance has also been proposed.\\
The propitiousness of deep learning has led researchers to study the effect of class imbalance in deep neural nets as well. Authors of \cite{buda2017systematic} study the effects of imbalance on convolutional neural networks and conclude that oversampling is the best-suited resampling technique and does not lead to overfitting in CNNs. Further motivating us to explore balancing data using oversampling.
Some other techniques to tackle class imbalance problem include the generation of new metrics \cite{boughorbel2017optimal} for optimization while working with class imbalanced data. One has been proposed in \cite{wang2016training}, the metrics proposed are based on mean  false positive and false negative error rates, and are used for optimization in neural networks and are better performing than mean squared error. We only discuss data level resampling (oversampling) techniques henceforth, which are befitting the scope of CBOS.\\
\section{Clustering Based Oversampling }
\label{sec3}
Several synthetic data resampling methods have had fair amount of success.
We propose a new data level oversampling method, Clustering Based Oversampling (CBOS) which has clustering as its basis. The success of techniques like SMOTE and ADASYN, which are predominantly based on kNN and also the sporadic success of randomly oversampling minority class motivated us to come up with a new data level oversampling technique. The proposed technique uses the euclidean distance in k-Means clustering ~\cite{kanungo2002efficient} to generate the cluster centroid and a distance normalization technique is used to generate the number of new data samples per existing minority class sample to be created. Although promising for low dimensional data space, the proposed algorithm is best suited for high dimensional data setting. This is due to the fact that with less number of attributes, the $Random$ parameter of our algorithm might sometimes generate similar samples leading to slight overfiting of the data.\\
Our algorithm does not take the distribution of majority class into consideration while oversampling the minority class samples and hence does not effect learning as done from majority class space. The main differences between this method and other algorithms like SMOTE and ADASYN are: first, the way this algorithm decides how many new samples are to be generated for each existing minority class sample; second, clustering helps to add the spatial structure of minority class into the new generated data samples; thirdly, the non-dependence of our technique on clustering technique used makes it more stable and finally, prevention of any change in the learning performance of the majority class. The proposed algorithm is explained below.\\
\hrule
\vspace{4pt}\textbf{Input.} Training Data having $ K $ samples,\\
\[ K = K_{l} + K_{m}  \quad (K_{m} >> K_{l})\]

where,  $ K_{l} $ is the number of samples belonging to the minority class, and, 
$ K_{m} $ is the number of samples belonging to the majority class.\\

$\textbf{1}.$ Assume the number of clusters in minority class data.\\

$\textbf{2}.$ Find the respective cluster centroid ($ C $) for each minority class sample
using k-Means clustering algorithm. \\

$\textbf{3}.$ For each $ x_{i}   \in   K_{l} $ calculate the euclidean distance ($ dist$) between the respective cluster centroid ($ C $) of $x_{i} $ and $ x_{i} $ itself, defined by,\\
\[ dist_{i} = Euclidean(x_{i},C) \]
$\textbf{4}.$ Normalize the distances over the whole data :\\
\hspace*{0.7cm}$ \quad \forall   x_{i} \in K_{l} $   \hspace*{0.7cm} do  \\
\[dist_{i}=\frac{dist_{i}}{\sum\limits_{i=1}^{K_{l}} dist_{i}}\]
$\textbf{5}.$ $ \forall  x_{i} \in K_{l} $ Calculate the number of samples ($n_{i}$) to be generated for each 

original minority class sample.
\[  n_{i}=integer\big(dist_{i} \times (K_{m}-K_{l})\times \eta\big) \] \\
here, $integer()$ is a function to round the resulting value, $K_{m}-K_{l}$ is the
difference between the number of samples of majority and the minority
class, 
$ \eta $ is the level of balance to be generated 1 meaning 100\%, 0.5 meaning 50\%
and so on. \\

$\textbf{6}.$ Generate  $n_{i}$ new samples for each $ x_{i} $ \\
\hspace*{0.8cm} \textbf{  for} $ i \in K_{l} $ \textbf{do} \\
\hspace*{1.2cm}  \textbf{ for} $ j \in n_{i} $ \textbf{do}  \\
\[d = abs(sample[i] - C)\]
\[newsample[i][j] = sample[i] \pm d\times Random(0,1) \]
\hspace*{1.2cm}  end \textbf{for}\\
\hspace*{0.8cm}  end \textbf{for}   	   	\vspace*{0.5cm}\\
$C$ is the respective cluster centroid vector of $x_{i}$, $ newsample[i][j]$ is the $j^{th}$ new sample vector for $x_{i}$ sample which has $ sample[i]$ as its original sample 
vector from minority class $K_{l}$ .\\

$abs()$ is the absolute element wise difference of the two vectors. $Random(0,1)$ is a random float between the $0,1$. For some cases the range can be reduced as well.\\
After generating all the new samples, their attribute values are reduced or increased to the original maximum and minimum values associated with each feature in original sample vector respectively. This is done to prevent generation of outliers. This step is also necessary to tackle the effect of overfitting as often seen in oversampling techniques.\\
CBOS can also be implemented by using a lower range for the $Random$ variable in step 5 of the algorithm, like ($0.2,0.4$) making range of $Random$ a very important parameter to be tuned for the proposed algorithm.\\

The idea behind this algorithm is that the more distant a sample point is from its cluster centroid, the lower the number of new points of this sample will be generated; and the less distant a sample point is from cluster centroid, the more is the number of new samples associated with this sample to be generated. This can be thought of as a centroid distance based weight metric, where the weights are used to find the number of new sample points to be generated.\\
The way our algorithm decides on the number of new samples for each original instance makes sure that the more important samples have higher representation than far lying samples in the new balanced minority class space. Also, as our algorithm does not take into consideration any effect of majority classs samples on minority samples, the classification accuracy of majority class is not affected at all. \\
\section{Learning Algorithm \&  Performance Metrics}
\label{sec4}
\subsection{Deep Neural Network}
\label{sec4a}
We have used Deep Neural Network (DNN) to learn the feature
representation from the data. DNN are becoming state-of-the-art as more data is being generated, its power lies in the fact that they work better on large datasets. The DNN used refers to an artificial neural network with more than one hidden layer(s). We have implemented DNNs with hidden layers ranging from 2 to 4, with different number of nodes in each layer and have reported the best results achieved for each resampling technique. 
\subsection{Assessment Metrics}
\label{sec4b}
Most of the classification performances are measured using accuracy or mean squared error. For imbalance class problem metrics taken from confusion matrix like F-score and the Area under ROC curve (AUC) \cite{fawcett2006introduction} have been quite often used.\\
The left column in the confusion matrix represents positive instances of the data set and the right column represents the negative ones. Therefore, the proportion of the two columns is representative of the class
distribution of the data set, and any metric that uses
values from both of these columns shall be inadvertently sensitive to
imbalances. Other accuracy measures namely, \em{precision, recall, F-score} \em have also been used for imbalance class problem. F-score is a weighted avergae of precision and recall, and has been used as one of the performance metrics in this work.
\[ F-score = (1+\beta^{2})\frac{Recall.Precision}{\beta^{2}.Recall+Precision}{} \]
$\beta$ is generally taken to be 1. 
To compare the performance of different classifier systems over a range of sample distributions, G-mean has also been calculated. \\
\[ G-Mean = \sqrt{\frac{TP}{TP+FN}\times\frac{TN}{TN+FP}}\]

\section{Experiments and Results}
\label{sec5}
In this section, we have experimentally shown the performance of different resampling techniques as compared to the proposed method.
\subsection{Experimental setting and data}
\label{sec5a}
All the  six datasets used for this study are binary classification data. The very famous MNIST hand written digits dataset has been used. From the MNIST \cite{lecun1998gradient} dataset, two classes are selected randomly and imbalance is induced in these resulting datasets. Specifically, four datasets comprising of the digit pairs (7, 5), (9, 8), (6, 2)  and (1, 4) have been used and an imbalance of 12\%, 10\%, 8\% and 6\% has been induced in them, respectively. These datasets are labeled as Data-1, Data-2, Data-3 and Data-4. \\The fifth dataset is a psychometric dataset from the SAPA \cite{978d379eca944e5581293c611919a87c} project called as SPI, extracted from R data repository and imbalance level is at 5\%. The SPI data has 145 attributes and it has been labeled as Data-5. A bi-classification auto insurance fraud dataset from UCI ML repository (represented as Data - 6) having an imbalance rate of 6\% and 32 atributes has also been used.\\

\vspace*{0.2cm}

\begin{table*}[h]
	\caption{Comparing the performance of resampling technqiues with different evaluation metrics}
	\centering
	\begin{tabular}{p{ 0.10\linewidth}p{0.14\linewidth}p{0.12\linewidth}p{0.10\linewidth}p{0.11\linewidth}p{0.10\linewidth}p{0.12\linewidth}}
		\hline	
		\textbf{Dataset} & \textbf{Algorithm} & \textbf{Precision} & \textbf{Recall} & \textbf{Accuracy} & \textbf{F-score} & \textbf{G-Mean} \Tstrut\Bstrut\\
		\hline 	\Bstrut
		& Imbalanced & 0.609 & 0.980 & 0.838 & 0.754 &  0.781 \Tstrut \Bstrut\\
		& SMOTE & 0.983 & 0.973 & 0.985 & \textbf{0.988} &  0.989 \\\Bstrut	\normalsize{Data-1} & SMT-ENN & 0.984 & 0.970 & 0.982 & 0.977 & 0.983 \\ \Bstrut
		& ADASYN & 0.987 & 0.975 & 0.984 & 0.981 &  0.986 \\
		& CBOS & \textbf{0.989} & \textbf{0.993} & \textbf{0.993} & 0.987 &  \textbf{0.992}  \Bstrut\\
		
		\hline
		& Imbalanced & 0.475 & 0.975 & 0.776 & 0.639 &  0.686 \Tstrut \Bstrut\\
		& SMOTE & 0.967 & 0.977 & 0.973 & 0.971 &  0.974 \\\Bstrut	\normalsize{Data-2} & SMT-ENN & 0.952 & \textbf{0.982} & 0.973
		& 0.964 & 0.970 \\ \Bstrut
		& ADASYN & 0.952 & 0.979 & 0.970 & 0.962 &  0.961 \Bstrut\\
		& CBOS & \textbf{0.972} & 0.977 & \textbf{0.981} & \textbf{0.977} &  \textbf{0.979} \Bstrut\\

		\hline
		& Imbalanced & 0.523 & 0.981 & 0.744 & 0.684 &  0.721 \Tstrut \Bstrut\\
		& SMOTE & 0.930 & 0.992 & 0.961 & 0.963 &  0.963 \\\Bstrut	\normalsize{Data-3} & SMT-ENN & 0.942 & 0.989 & 0.965
		& 0.964 & 0.964 \\ \Bstrut
		& ADASYN & 0.936 & \textbf{1.000} & \textbf{0.972} & 0.963 &  0.966\Bstrut\\
		& CBOS & \textbf{0.947} & 0.997 & 0.971 & \textbf{0.973} &  \textbf{0.969} \Bstrut\\
		\hline
		& Imbalanced & 0.484 & 0.941 & 0.704 & 0.624 &  0.683 \Tstrut \Bstrut\\
		& SMOTE & 0.910 & 0.971 & 0.932 & 0.933 &  0.928 \\\Bstrut	\normalsize{Data-4} & SMT-ENN & 0.902 & 0.979 & 0.927
		& 0.924 & 0.919 \\ \Bstrut
		& ADASYN & 0.916 & \textbf{0.971} & 0.938 & 0.937 &  0.931 \Bstrut\\
		& CBOS & \textbf{0.922} & 0.970 & \textbf{0.941} & \textbf{0.943} &  \textbf{0.939} \Bstrut\\
		\hline
		& Imbalanced & 0.471 & 0.794 & 0.784 & 0.596 &  0.668 \Tstrut \Bstrut\\
		& SMOTE & 0.867 & 0.851 & 0.901 & 0.859 &  0.908 \\\Bstrut	\normalsize{Data-5} & SMT-ENN & 0.840 & 0.761 & 0.852
		& 0.794 & 0.849 \\ \Bstrut
		& ADASYN & \textbf{0.903} & 0.937 & \textbf{0.948} & 0.919 &  \textbf{0.936} \Bstrut\\
		& CBOS & 0.892 & \textbf{0.960} & \textbf{0.948} & \textbf{0.920} &  0.929 \Bstrut\\
		\hline
		& Imbalanced & 0.538 & 0.841 & 0.824 & 0.628 &  0.782 \Tstrut \Bstrut\\
		& SMOTE & 0.931 & \textbf{0.901} & 0.901 & \textbf{0.918} &  0.915 \\\Bstrut	\normalsize{Data-6} & SMT-ENN & 0.914 & 0.891 & 0.908
		& 0.903 & 0.912 \\ \Bstrut
		& ADASYN & \textbf{0.943} & 0.889 & \textbf{0.919} & 0.913 &  \textbf{0.941} \Bstrut\\& CBOS & 0.906 & 0.897 & 0.902 & 0.901 &  0.914 \Bstrut\\
		\hline
		\label{tab:3}
	\end{tabular}
\end{table*}
Based on the performance metrics discussed in Section \ref{sec4b}, Table \ref{tab:3}  presents the comparison between the proposed CBOS and other resampling techniques. The results shown are averaged over 10 runs. The resampling algorithms are then applied on the training data and tested on the test set. Best parameters for SMOTE, SMOTE-ENN and ADASYN have been selected. The DNN has been tuned to get the best peformance for all resampling techniques similarly.\\
\subsection{Analysing the Performance}
\label{sec5b}
From Table \ref{tab:3}, it can be easily inferred that the proposed method CBOS is generally the best across all test benches. Performance results from the original imbalanced dataset have also been included and all the resampling methods improve upon the original imbalanced data. The original data show good performance in certain metrics recall. This is because only one class is imbalanced and directly using a learning classifier on original data results in good classification of majority class only. \\
As the level of imbalance becomes severe from Data-1 to Data-4, we see fall in the value of mostly all the metrics. This implies that more severe the imbalance, the less effective the resmpling techniques are. Comparing the results of resampling techniques with respect to the majority class in Data-1, we see that the Recall score of all the techniques sans CBOS decreases as compared to the original imbalanced data. This shows that the way in which CBOS resamples minority class data does not in any way hinder in classification of the majority class whereas other techniques in some way have an effect on the majority class data in addition to the minority class data. The best performing technique for each metric has been highlighted, and the most highlighted across all datasets is the CBOS technique.\\
On comparing the performance results of Data-5 and 6, we see that the attribute to sample ratio of Data-5 is very large than Data-6. This results in an increase in performance of resampling techniques as SMOTE and ADASYN, but, CBOS also produces results in a tantamount range. This leads us to say that SMOTE, ADASYN and other similar resampling techniques perform better in low dimesnional setting. The proposed method, CBOS performs comparable to others for low dimensional data but is more efficient in case of high dimensional data space. 
\subsection{Discussions}
\label{sec5c}
The strength of CBOS lies in the fact that in addition to the oversampling minority classs accurately, CBOS produced new samples do not effect majority class space in any way. We use the randomness in an effective way by restraining the maximum and minimum values of the newly genrated samples. Also rather than using kNN for generation of new data points we use cluster centroids. This means our proposed algorithm also incorporates the distribution structure of the minority class data in the newly resampled data, not commonly seen in other synthetic resampling techniques. The proposed algorithm is more suited for medium and high dimensional data because of the $Random$ function used in the algorithm. If the need of $Random$ for CBOS is reduced in future, CBOS can be made adaptable to very low dimensional data as well. \\
The results shown are only for deep neural networks but CBOS algorithm can be used with other learning algorithms as well. Ensemble models have proven to be great classifier systems and in future CBOS can be integrated with ensemble models to further enhance the classification prowess of CBOS. As most of the imabalance problems are dominated by two class classification, this work only studies CBOS for two class classification problem but CBOS can also be used for multi class imbalance problems. More imbalanced datasets can be used to test the efficacy of the proposed algorithm. We have shown CBOS when used with k-Means clustering but CBOS's results when used with other clustering algorithms are expected to be in comparable ranges. In future, CBOS can be implemented with other clustering models for example Mean-shift clustering or the recently introduced, expectation maximization clustering using Gaussian mixture models \cite{reynolds2015gaussian}. An algorithm is considered novel and robust when it has lower number of parameters it requires to be tuned on, in this regard, we would like to reduce the efffect that $Random$ parameter has in generating new data samples, making CBOS more stable and robust to varied type of data.

\section{Conclusion}
\label{sec6}
A lot of work has been done for solving class imbalance problem using data resampling, but most of these methods in some way effect the majority class space as well. CBOS method prevents generation of outliers and it does not affect majority class set in any way as CBOS's oversampling technique does not take interdependence of the two classes into consideration. CBOS uses a completely different way to decide on the number of new samples to be generated for each existing minority class sample. CBOS uses clustering and cluster centroids for developing new data points meaning overall distribution of minority class is also taken into consideration. The distance normalization technique used for calculation of the number of new sample points to be generated is designed in a way that it lets the data points nearer to centroid have more say in the generation of new sample points. Motivated by the results in this paper, we believe that CBOS might prove a powerful method in multi class imbalanced learning as well.

\bibliographystyle{abbrv}
\bibliography{bibliography}
\small

\end{document}